%% file: main.tex
\documentclass[10pt,twocolumn,letterpaper]{article}

\usepackage[pagenumbers]{cvpr} 

\input{preamble}
\definecolor{cvprblue}{rgb}{0.21,0.49,0.74}
\usepackage[pagebackref,breaklinks,colorlinks,allcolors=cvprblue]{hyperref}

\def\confName{CVPR}
\def\confYear{2027}

\title{Trailblazer-F: Speculative Action-Flow Inference for Autonomous Driving}

\author{
    Haibo Hu$^{1}$\quad Lianming Huang\textsuperscript{\rm 1}\thanks{Corresponding author:lmhuang8-c@my.cityu.edu.hk}\quad Qiao Li\textsuperscript{\rm 2}\quad  Nan Guan$^{1}$\quad  Chun
    Jason Xue$^{2}$ \\
    $^{1}$City University of Hong Kong\\
    $^{2}$Mohamed bin Zayed University of Artificial Intelligence
}

\begin{document}
\maketitle
\input{sec/sections/00_abstract}

\input{sec/sections/01_introduction}
\input{sec/sections/02_related_work}
\input{sec/sections/motivation}
\input{sec/sections/03_method}
\input{sec/sections/04_experiments}

\input{sec/sections/05_conclusion}

\input{sec/sections/06_llm_usage}
{
    \small
    \bibliographystyle{ieeenat_fullname}
    \bibliography{main}
}


\end{document}

%% file: sec/sections/00_abstract.tex
\begin{abstract}
Driving world-action models generate planned trajectories through iterative action denoising, with each update depending on the preceding state. This sequential dependency incurs substantial inference latency, limiting timely planning. Speculative inference offers an opportunity to reduce this latency by predicting future denoising states for parallel target evaluation, but requires efficient proposal generation and suitable verification for continuous actions. We present Trailblazer-F, a speculative action-flow inference framework in which a lightweight drafter acts as a trailblazer, charting a candidate denoising path through the compact, structured waypoint space so that the target model can verify multiple steps in parallel and advance with fewer sequential updates.
Our method maintains planning quality while keeping the pretrained target model frozen.
Across all 12,146 NAVSIM v1 navtest scenes, Trailblazer-F
maintains near-baseline planning performance, with PDMS changes
of $+0.04$ and $-0.08$ points for SimWAM and DriveWAM, respectively.
Paired inference measurements on an NVIDIA L20 GPU yield
$2.31\times$ and $1.55\times$ speedups.
Real-vehicle experiments on NVIDIA Jetson AGX Orin further
demonstrate $1.96\times$ and $1.37\times$ speedups, together
with close qualitative trajectory agreement with native sampling. our code is here 
\url{https://github.com/DanileHaibo/Trailblazer-F}
\end{abstract}

%% file: sec/sections/01_introduction.tex
\section{Introduction}
\label{sec:introduction}

\input{sec/sections/01_intro_paragraph2}

\input{sec/sections/01_intro_paragraph3}

\input{sec/sections/01_intro_paragraph4}
Motivated by these challenges and observations, we propose
\textbf{Trailblazer-F}, a speculative inference framework tailored to
action flow in autonomous-driving world models. Like a pathfinder
scouting the route ahead, a lightweight drafter anticipates future
denoising states, allowing the target model to verify multiple proposals
in parallel. The drafter exploits the compact action representation
to predict span-specific endpoints in a single forward pass, then
constructs future states analytically using the known noise schedule.
For verification, the target independently computes solver-aligned,
one-step corrections from the corresponding proposed predecessor states.
Each candidate must satisfy a normalized discrepancy threshold at every
waypoint, and only the consecutive passing prefix is accepted.
Together, these designs exploit the structure of action flow to reduce
sequential inference latency while controlling local deviations from
the target dynamics, as shown in Fig.~\ref{fig:introduction}(b).

\paragraph{Contributions:}
\begin{itemize}
  \item We introduce Trailblazer-F, to the best of our knowledge
  the first speculative decoding framework specifically designed
  to accelerate action-flow denoising in autonomous-driving
  world-action models.

  \item We develop three complementary components: a lightweight
  action-flow drafter for multi-step state prediction,
  state-based parallel verification that independently generates
  solver-aligned target corrections, and a physically interpretable
  acceptance mechanism that checks normalized position and heading
  consistency at every waypoint.

  \item We evaluate Trailblazer-F on all 12,146 NAVSIM v1
  \texttt{navtest} scenes across two world-action models,
  maintaining near-baseline planning scores while achieving
  up to $2.32\times$ inference speedup on an NVIDIA L20 GPU.
  Real-vehicle experiments on NVIDIA Jetson AGX Orin further
  demonstrate onboard acceleration and qualitative trajectory
  agreement with native sampling.
\end{itemize}

%% file: sec/sections/01_intro_paragraph2.tex
\begin{figure}[!t]
\centering
\includegraphics[width=0.5\textwidth, trim=0 2cm 0cm 0cm, clip]{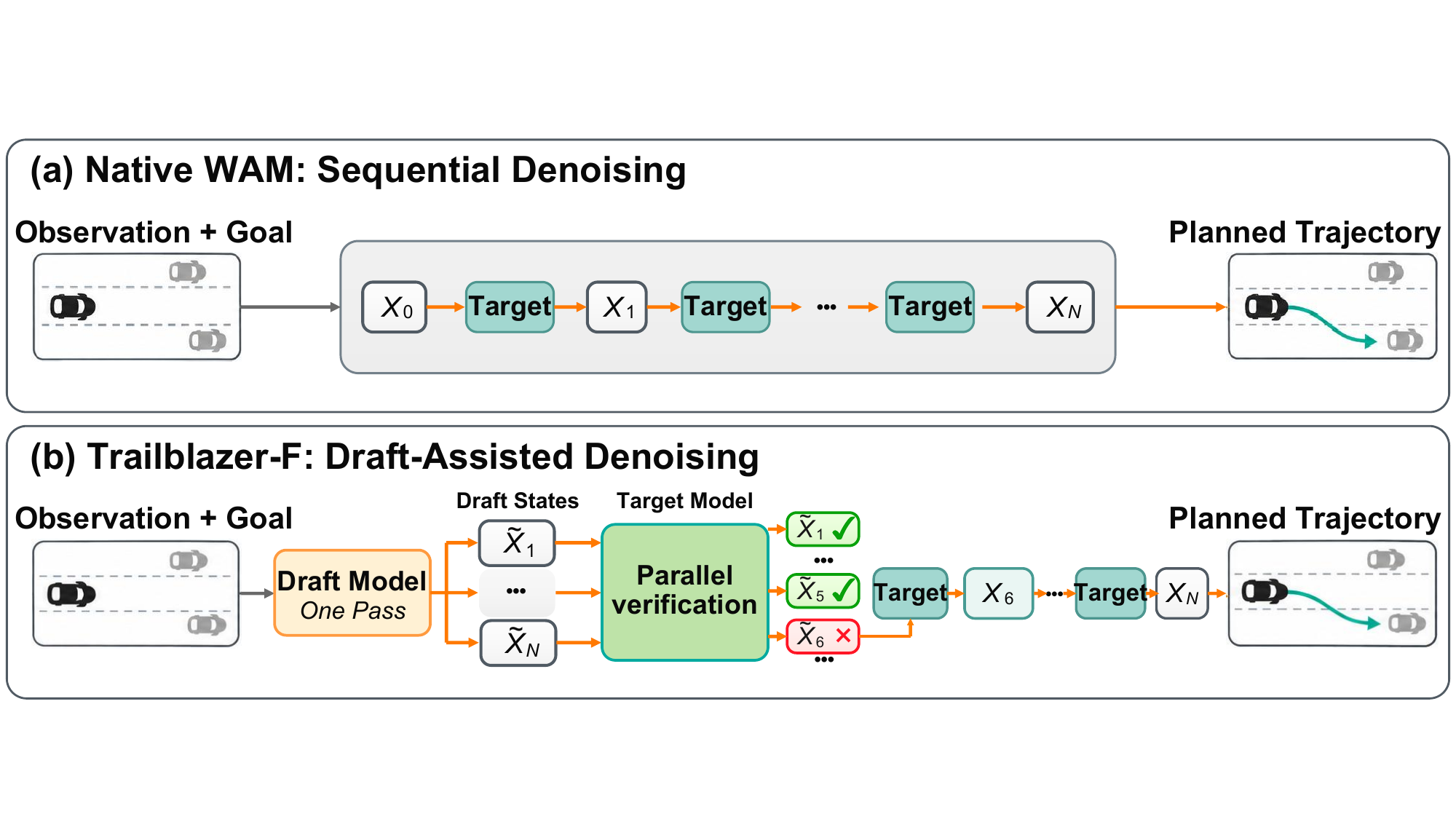}
\caption{Action Flow in World Model architectures and Speculative Action-Flow inference.}
\label{fig:introduction}
\end{figure}

World models are emerging as a promising paradigm for autonomous
driving, connecting future environment prediction with action
evaluation and planning~\cite{wang2024drivedreamer,gao2024vista,li2025law}.
Flow matching has become widely adopted in recent vision-based
world-action models for learning visual dynamics and continuous
driving actions~\cite{zhou2026drivedreamerpolicy,shi2026drivewam,zhao2026simwam,wang2026svwam,parikh2026physwam}.
However, standard sampling introduces a sequential bottleneck:
each denoising update requires a target-model evaluation
conditioned on the preceding state, as shown in
Fig.~\ref{fig:introduction}(a)~\cite{lipman2023flow,shi2026drivewam,wang2026svwam}.
Our profiling shows that action denoising accounts for most
inference latency in the evaluated models.
This overhead conflicts with the need for timely planning
in dynamic traffic~\cite{jiang2023vad,liao2025diffusiondrive}.
Reducing the cost of action denoising while preserving planning
quality is therefore a key challenge in bringing world models
to real-time autonomous driving.

%% file: sec/sections/01_intro_paragraph3.tex
Existing efforts to accelerate generative inference include
distilling multi-step samplers into fewer-step
generators~\cite{salimans2022progressive}, learning truncated
denoising processes for driving
policies~\cite{liao2025diffusiondrive}, and designing action
branches that avoid future-video generation at
inference~\cite{zhao2026simwam,wang2026svwam}.
These approaches improve efficiency by modifying the model
architecture or changing the sampling dynamics, which can alter
the generator's output distribution.
Consequently, the planning behavior of the original pretrained
model is not necessarily preserved.
This motivates a complementary objective: accelerating action-flow
inference while keeping the pretrained target model's parameters
and architecture unchanged and preserving its planning quality.

Speculative decoding offers a promising approach to this objective:
a lightweight drafter proposes multiple candidates for parallel
target verification, reducing sequential evaluations while retaining
the pretrained target~\cite{leviathan2023speculative,chen2023speculative,li2024eagle}.
Despite progress in diffusion and flow
generation~\cite{debortoli2025speculative,bajpai2026flowcast},
applying this paradigm to world models for autonomous driving
requires overcoming two key challenges.
First, \emph{inexpensive multi-step state prediction}: future
denoising states depend on successive target updates, making it
difficult for a lightweight drafter to predict several states
accurately without reproducing the costly sequential computation.
Second, \emph{reliable continuous-state verification}: exact
matching is too restrictive for continuous states, whereas
accepting inaccurate proposals can degrade planning quality.
An effective verifier must therefore distinguish useful
approximations from unacceptable deviations from the target dynamics.

By unfolding the ten-step action-flow denoising process in
SimWAM and DriveWAM and inspecting its intermediate states,
we observe two properties that help address these challenges.
First, \emph{a compact waypoint representation}: each state
contains eight waypoints with position and heading coordinates,
forming an $8\times3$ representation throughout
denoising~\cite{zhao2026simwam,shi2026drivewam}.
This low-dimensional structure motivates lightweight prediction
of multiple future denoising states.
Second, \emph{consistent physical semantics}: the state coordinates
remain associated with waypoint positions and headings across
denoising steps, even when intermediate states are
noisy~\cite{shi2026drivewam,zhou2026drivedreamerpolicy,wang2026svwam}.
A candidate state and its target correction at the same denoising
time can therefore be compared waypoint by waypoint through aligned
position and heading channels in a common normalized coordinate system.
This correspondence provides an interpretable basis for
tolerance-based verification against the target dynamics.

%% file: sec/sections/02_related_work.tex
\section{Related Work}
\label{sec:related}

\paragraph{Driving world-action models.}
Driving world-action models (WAMs) connect future-scene prediction with
ego-motion generation, transferring generative world representations
into driving policies. Epona~\cite{zhang2025epona} employs rectified flow
for both next-frame and trajectory generation, conditioned on shared
spatiotemporal features. DriveVA~\cite{liu2026driveva} jointly generates
future video latents and action tokens through a shared flow-matching
decoder, strengthening the connection between imagined scene evolution
and planned motion. DriveWAM~\cite{shi2026drivewam} similarly adapts
pretrained video generative priors to an autoregressive video-action
policy with a joint flow-matching objective. Complementary designs
introduce specialized experts or additional geometric supervision.
SimWAM~\cite{zhao2026simwam} co-trains video and action experts through
joint flow matching, while isolating their attention so that action
inference does not require future-video generation.
DriveDreamer-Policy~\cite{zhou2026drivedreamerpolicy} incorporates
geometry-aware conditioning through modular depth, video, and action
generators. PhysWAM~\cite{parikh2026physwam} jointly denoises multiview
video, metric depth, and ego motion within a single flow-matching
transformer, with geometric constraints coupling their predictions.
Together, these developments establish action flow as a widely adopted
formulation for trajectory generation in recent driving WAMs. Despite
differences in architecture and conditioning, these models generate
actions by integrating learned velocity fields, making the efficiency
of iterative action refinement an important concern for deployment.
\begin{figure*}[!t]
\centering
\includegraphics[width=1\textwidth, trim=0 6.5cm 0cm 6cm, clip]{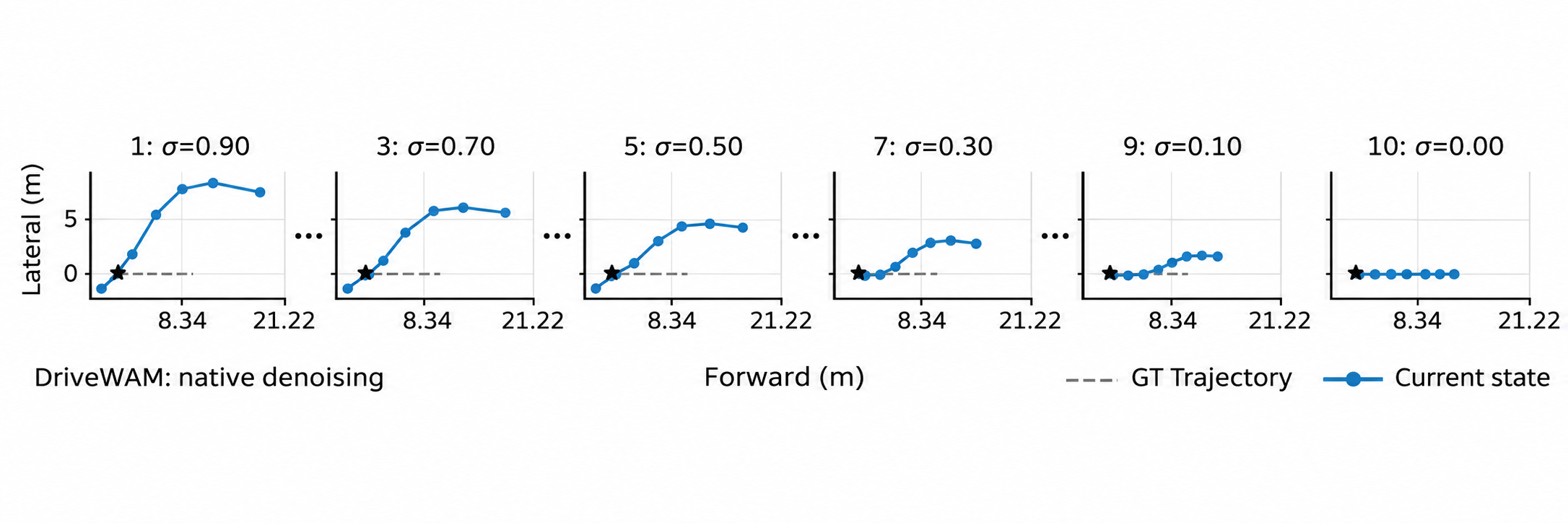}
\caption{\textbf{Action-flow denoising in DriveWAM.}
Six of the ten native sampler steps as the noise level $\sigma$
decreases: blue is the current state, gray dashed the reference
trajectory, black star the ego origin. Each state is a compact
$8\times3$ encoding of waypoint positions and headings; ellipses
mark omitted steps.}
\label{fig:motivation}
\end{figure*}

\paragraph{Accelerating action-flow inference.}
Existing efforts improve inference efficiency through changes to
architecture, training, or sampling. Architectural decoupling, as in
SimWAM~\cite{zhao2026simwam}, removes future-video synthesis from the
planning path but retains iterative action denoising. In related
generative driving planners, DiffusionDrive~\cite{liao2025diffusiondrive}
combines trajectory anchors with a truncated diffusion formulation
to support few-step planning. More directly within WAMs,
OSWAM~\cite{zhang2026oswam} enables one-step action generation through
learned structured initialization and privileged-information
distillation. Such designs can retain competitive planning performance,
but require specialized training or changes to the action-generation
architecture. Conversely, aggressive step reduction without corresponding
adaptation can degrade planning quality, as illustrated by the
one-step baselines reported in OSWAM~\cite{zhang2026oswam}.

Speculative inference offers a complementary approach based on target
verification. Following speculative decoding~\cite{leviathan2023speculative},
De Bortoli et al.~\cite{debortoli2025speculative} preserve target
distributions for stochastic diffusion chains, but their procedure does
not directly apply to deterministic samplers.
FlowCast~\cite{bajpai2026flowcast} uses velocity extrapolation and parallel
error-based verification for visual generation. Closely related,
Realtime-VLA FLASH~\cite{niu2026realtimevlaflash} drafts action chunks, reuses
context across replanning rounds, and checks the consistency of reconstructed endpoints to select executable action prefixes.
Trailblazer-F instead accelerates denoising within each WAM planning
call: a lightweight drafter predicts scene-conditioned, multi-span
endpoints and reconstructs candidate states using the current state and
noise schedule. Parallel verification compares each candidate with its
native target update through physically interpretable waypoint-wise
position and heading discrepancies. Acceptance advances along denoising
steps rather than action execution, retaining the target architecture,
parameters, and sampling schedule.

%% file: sec/sections/motivation.tex
\section{Motivation}
\paragraph{Motivation 1: Learnable denoising in a compact action space.}
As \cref{fig:motivation} illustrates, action denoising follows a
progressive refinement process, with waypoint positions gradually
approaching the final trajectory.
This structured evolution suggests that future states exhibit
regularities that a lightweight predictor can learn.
Moreover, each action state in the evaluated WAMs contains only
$8\times3$ position and heading coordinates, making the prediction
problem compact.
These observations motivate collecting intermediate denoising-state
sequences from the frozen target and using them to train a small MLP.
By learning the evolution of these low-dimensional states, the
drafter can propose multiple future states in a single forward pass.

\paragraph{Motivation 2: Physically interpretable state comparisons.}
Action-flow states preserve explicit position and heading semantics
throughout denoising, even when intermediate states remain noisy.
At the same denoising time, a candidate and its target correction
can therefore be compared waypoint by waypoint in a shared normalized
coordinate system.
We introduce a tolerance to avoid rejecting useful candidates solely
because of minor discrepancies, with acceptance requiring the discrepancy
at every waypoint to remain below the threshold.
The physical meaning of the coordinates also makes verification
interpretable: the acceptance decision reflects deviations in waypoint
positions and headings, providing a concrete geometric interpretation
of local consistency with the target model.

%% file: sec/sections/03_method.tex
\section{Trailblazer-F}
\label{sec:method}

\begin{figure*}[!t]
\centering
\includegraphics[width=1\textwidth, trim=0 2cm 0cm 3cm, clip]{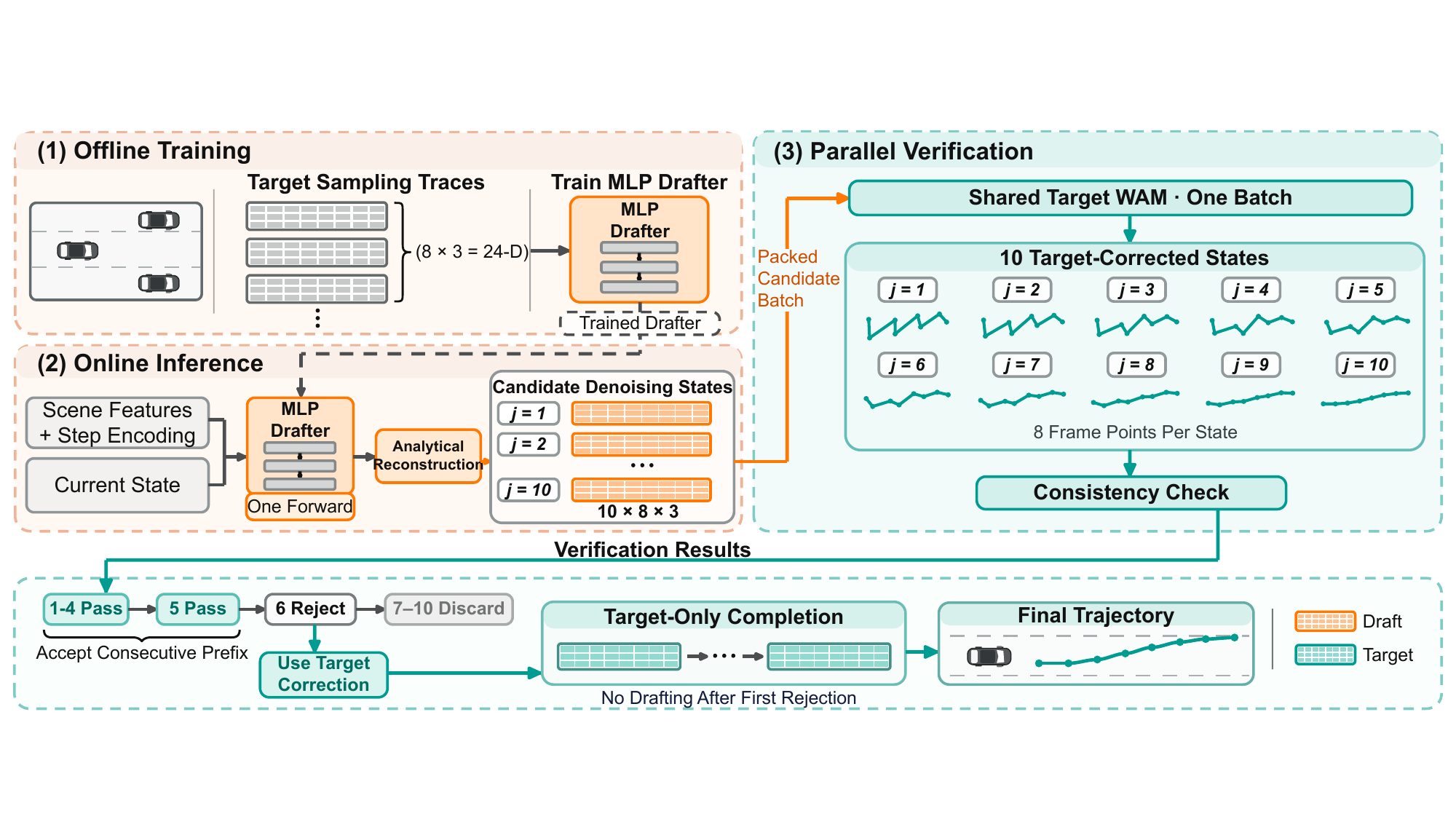}
\caption{\textbf{Overview of Trailblazer-F.}
A lightweight drafter learns from frozen-target denoising traces
and proposes multiple states through analytical reconstruction.
Parallel target verification and waypoint-wise consistency checks
determine the accepted draft prefix.
At the first rejection, the sampler adopts the target-corrected
state, discards subsequent candidates, and completes denoising
with the target alone.}
\label{fig:overview}
\end{figure*}

Figure~\ref{fig:overview} illustrates the training and inference
pipeline of Trailblazer-F.
A lightweight MLP drafter is trained offline on denoising traces
collected from the frozen target model.
During inference, a single draft forward pass predicts multi-span
endpoints, which are analytically reconstructed into candidate
denoising states.
The target then evaluates their corresponding predecessor states
in parallel to obtain single-step corrections.
Waypoint-wise consistency checks determine acceptance, and only
the consecutive passing prefix of draft states is retained.
At the first rejection, the target-corrected state is adopted,
all remaining candidates are discarded, and sampling continues
with the target alone.
\subsection{Problem Formulation}
\label{sec:problem_formulation}

Given scene conditions $c$, an initial noisy action state
$X_0\in\mathbb{R}^{8\times3}$, and a pretrained target model $\theta$,
we aim to generate an eight-waypoint ego trajectory with lower
inference latency while preserving planning performance.
The serial reference performs sequential denoising along a fixed
schedule $\mathcal{S}=(s_0,\ldots,s_N)$. We retain both the target
weights and this schedule.

Let $\mathfrak{T}(\theta,\mathcal{S})$ denote admissible accelerated
sampling procedures. 
We use $\mathcal{D}_t$ and $\mathcal{D}_q$ as the evaluation sets for inference latency and planning performance, respectively.
For $\mathcal{T}\in\mathfrak{T}$, let
$L(\mathcal{T})$ denote mean inference latency on $\mathcal{D}_t$
and $m_j(\mathcal{T})$ the $j$-th planning metric on $\mathcal{D}_q$,
measured under a fixed protocol $\Pi$. With metrics oriented so
that higher values are better, our objective is
\begin{equation}
\begin{aligned}
\min_{\mathcal{T}\in\mathfrak{T}(\theta,\mathcal{S})}
\quad & L(\mathcal{T}) \\
\text{s.t.}\quad
& m_j(\mathcal{T})
\ge m_j(\mathcal{T}_{\mathrm{serial}})-\epsilon_j,\\
& j=1,\ldots,M,
\end{aligned}
\label{eq:overall_formulation}
\end{equation}
where $T_{\mathrm{serial}}$ denotes the native sequential sampler,
$M$ is the number of planning metrics, and $\epsilon_j\ge0$ specifies the acceptable degradation
relative to the serial reference.

Our procedure $\mathcal{T}_{\theta,\phi}[\mathcal{S},\tau]$
uses a lightweight drafter $\phi$,
and local verification threshold $\tau$, distinct from $\epsilon_j$.
For example, if five draft states pass verification and the sixth
is rejected, we retain the accepted prefix, adopt the target-corrected
$X_6$, and complete denoising to $X_N$ with the target.
Latency includes drafting, verification, and target completion.

\subsection{Action-Flow Drafting and Training}
\label{sec:draft_training}

\begin{figure}[!t]
\centering
\includegraphics[width=0.5\textwidth, trim=8cm 1cm 6cm 0cm, clip]{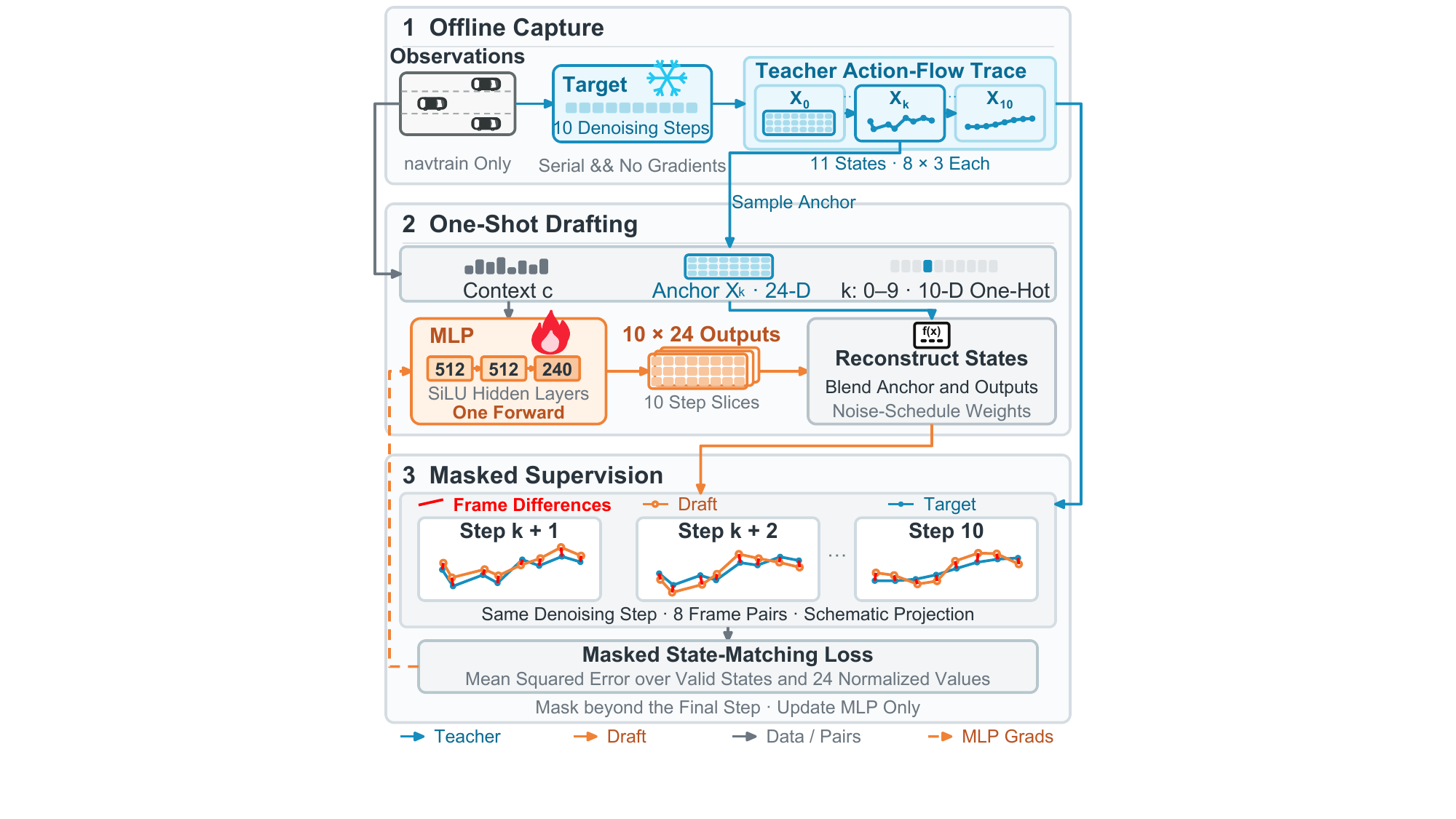}
\caption{Offline training of the action-flow drafter}
\label{fig:train}
\end{figure}

As shown in \cref{fig:overview}, Trailblazer-F begins with
offline training of a lightweight action-flow drafter.
The drafter learns to predict multiple future denoising states
from the current state and scene context.
\cref{fig:train} details this process, from collecting
target sampling traces to constructing supervised predictions.

We first execute the frozen target on \texttt{navtrain} scenes
using its native $N=10$ denoising steps.
For each scene, we retain the context features $c$ and the complete
normalized action-state sequence
$\{X_k^{\mathrm{T}}\}_{k=0}^{N}$, including the initial noisy state.
Each trace therefore contains 11 states, each represented as an
$8\times3$ array of waypoint positions and headings.
Different target models provide their own sampling traces and
are paired with separately trained drafters.

To construct a training example, we select an anchor step
$k\in\{0,\ldots,N-1\}$ and set $X_k=X_k^{\mathrm{T}}$.
The drafter takes the scene context $c$, the flattened anchor
state, and a ten-dimensional one-hot step encoding $e(k)$.
It consists of two 512-unit hidden layers with SiLU activations
and a 240-dimensional output layer.
A single forward pass produces $N$ span-specific endpoint
estimates, each containing 24 coordinates:
\begin{equation}
    \bigl(\widehat{A}_{k,1},\ldots,\widehat{A}_{k,N}\bigr)
    = D_{\phi}\bigl(c,X_k,e(k)\bigr),
    \label{eq:draft_prediction}
\end{equation}
where $\widehat{A}_{k,j}\in\mathbb{R}^{8\times3}$ parameterizes
a proposal $j$ denoising steps beyond the anchor.

These estimates are converted into candidate states using the
known noise schedule:
\begin{equation}
    \widetilde{X}_{k+j}
    =
    \frac{\sigma_{k+j}}{\sigma_k}X_k
    +
    \left(1-\frac{\sigma_{k+j}}{\sigma_k}\right)
    \widehat{A}_{k,j},
    \label{eq:candidate_reconstruction}
\end{equation}
for $j=1,\ldots,K_k$, where $K_k=N-k$ and $\sigma_k>0$.
All candidates share the same anchor and are constructed without
recursive draft-model calls.
This reconstruction remains within the training computation
graph, allowing supervision on future states to update the
endpoint estimates.

As illustrated by the masked supervision in
\cref{fig:train}, each candidate is paired with the
recorded target state at the same denoising time.
Predictions extending beyond the final step are masked out.
For a batch of $B$ training examples with anchor steps $k_b$,
the training objective is
\begin{equation}
    \mathcal{L}_{\mathrm{draft}}
    =
    \frac{
        \displaystyle\sum_{b=1}^{B}
        \sum_{j=1}^{K_{k_b}}
        \left\|
        \widetilde{X}^{(b)}_{k_b+j}
        -
        X^{\mathrm{T},(b)}_{k_b+j}
        \right\|_F^2
    }{
        \displaystyle 24\sum_{b=1}^{B}K_{k_b}
    }.
    \label{eq:draft_training_loss}
\end{equation}
The loss averages squared errors over all valid future states
and their normalized coordinates.
Only the drafter parameters $\phi$ are optimized; the target
remains frozen.
The endpoint estimates thus receive supervision through the
future states they reconstruct.



\subsection{State-Based Parallel Verification}
\label{sec:parallel_verification}

During online inference, the trained drafter constructs a block
of candidate states
$\{\widetilde{X}_{k+j}\}_{j=1}^{K_k}$
using \cref{eq:draft_prediction,eq:candidate_reconstruction}.
As shown in \cref{fig:overview}, these candidates provide
the intermediate inputs needed to evaluate multiple target
updates in parallel.
For candidate $\widetilde{X}_{k+j}$, verification starts from
its corresponding predecessor $\widetilde{X}_{k+j-1}$,
with $\widetilde{X}_k=X_k$.
We stack these predecessors along the batch dimension:
\begin{equation}
    \mathcal{X}_k
    =
    \operatorname{Stack}
    \bigl(\widetilde{X}_k,\ldots,
    \widetilde{X}_{k+K_k-1}\bigr)
    \in\mathbb{R}^{K_k\times8\times3}.
    \label{eq:verification_batch}
\end{equation}
Each batch entry is paired with its own denoising time
$t_{k+j-1}$ and the same scene context $c$.
\begin{table*}[t]
  \centering
  \caption{\textbf{Planning performance and inference efficiency on NAVSIM v1.} Camera inputs: SV (single-view), MV (multi-view).}
  \label{tab:local_results}
  \small
  \setlength{\tabcolsep}{2pt}
  \begin{tabular*}{\textwidth}{@{\extracolsep{\fill}}lcrrrrrrrrr@{}}
    \toprule
    Method & Sensors & NC $\uparrow$ & DAC $\uparrow$ & TTC $\uparrow$
    & C. $\uparrow$ & EP $\uparrow$ & PDMS $\uparrow$
    & \shortstack{Inference\\(ms) $\downarrow$} & Speedup $\uparrow$
    & \shortstack{Accept.\\(steps/10) $\uparrow$} \\
    \midrule
    \multicolumn{11}{@{}l}{\textit{Selected planning methods}} \\
    Human
    & -- & 100.0 & 100.0 & 100.0 & 99.9 & 87.5 & 94.8
    & -- & -- & -- \\
    UniAD~\cite{hu2023uniad}
    & MV & 97.8 & 91.9 & 92.9 & 100.0 & 78.8 & 83.4
    & 555.6 & -- & -- \\
    Hydra-MDP++ (R34)~\cite{li2025hydramdppp}
    & MV & 97.6 & 96.0 & 93.1 & 100.0 & 80.4 & 86.6
    & 206.2 & -- & -- \\
    \midrule
    \multicolumn{11}{@{}l}{\textit{VLA-based methods}} \\
    ReCogDrive~\cite{li2026recogdrive}
    & MV & 98.1 & 94.7 & 94.2 & 100.0 & 80.9 & 86.5
    & 306.1 & -- & -- \\
    AutoVLA~\cite{zhou2025autovla}
    & MV & 98.4 & 95.6 & 98.0 & 99.9 & 81.9 & 89.1
    & 1310.0 & -- & -- \\
    \midrule
    \multicolumn{11}{@{}l}{\textit{World-action methods }} \\
    PWM~\cite{zhao2025pwm}
    & SV & 98.6 & 95.9 & 95.4 & 100.0 & 81.8 & 88.1
    & 850.0 & -- & -- \\
    SV-WAM~\cite{wang2026svwam}
    & MV & 98.6 & 98.6 & 95.1 & 100.0 & 83.9 & 90.2
    & 341.6 & -- & -- \\
    \midrule
    \multicolumn{11}{@{}l}{\textit{SimWAM}} \\
    Native~\cite{zhao2026simwam}
    & SV & 97.77 & 94.11 & 93.69 & 100.00 & 81.50 & 86.04
    & 460.3 & $1.00\times$ & -- \\
    \quad + Trailblazer-F
    & SV & 97.76 & 94.22 & 93.57 & 99.99 & 81.58 & 86.08
    & \textbf{198.9} & $\mathbf{2.31\times}$ & 6.8 \\
    \midrule
    \multicolumn{11}{@{}l}{\textit{DriveWAM}} \\
     Native~\cite{shi2026drivewam}
    & SV & 98.33 & 98.15 & 95.40 & 99.99 & 84.33 & 90.25
    & 1015.9 & $1.00\times$ & -- \\
    \quad + Trailblazer-F
    & SV & 98.31 & 98.11 & 95.32 & 99.99 & 84.28 & 90.17
    & \textbf{655.2} & $\mathbf{1.55\times}$ & 7.0 \\
    \bottomrule
  \end{tabular*}
\end{table*}

The frozen target processes this batch through its original
action projection, time conditioning, and Transformer layers.
Different denoising stages occupy separate batch entries,
without attention across entries, while each entry preserves
the target's original action-token interactions and access
to scene context.
One batched forward pass produces $K_k$ velocity fields,
which yield target-corrected states through the native
Euler update:
\begin{equation}
    \begin{aligned}
        \overline{X}_{k+j}
        &=
        \widetilde{X}_{k+j-1} \\
        &\quad+
        \Delta t_{k+j-1}\,
        v_{\theta}\bigl(
        \widetilde{X}_{k+j-1},t_{k+j-1};c
        \bigr),
    \end{aligned}
    \label{eq:parallel_target_correction}
\end{equation}
where $j=1,\ldots,K_k$, $t_i$ denotes the target's native
solver time, and $\Delta t_i=t_{i+1}-t_i$.

Each correction is computed from its proposed predecessor,
without waiting for another correction to finish.
The parallel unit is therefore a complete action-state
transition, rather than a token position within a shared
sequence.
These updates assess local consistency with the target
dynamics; they do not reproduce a full serial target rollout.
The resulting pairs
$(\widetilde{X}_{k+j},\overline{X}_{k+j})$
represent proposed and corrected states at the same denoising
time and are passed to the waypoint-wise acceptance mechanism.

\subsection{Acceptance Based on Physical Attributes}
\label{sec:physical_acceptance}
    
Action-flow states retain an explicit correspondence to physical
attributes: each row represents the position and heading of a
particular waypoint. This correspondence remains available
throughout denoising, even when intermediate states contain noise.
We exploit this structure to assess target consistency at every
waypoint. A single error averaged over the entire $8\times3$
    state can obscure a large discrepancy at an individual waypoint.
    We therefore impose the acceptance tolerance separately at
    each waypoint.
    
    For candidate $j$ and waypoint $r$, we compute the root-mean-square
    discrepancy across the normalized position and heading channels:
    \begin{equation}
        d_{j,r}
        =
        \sqrt{
        \frac{1}{3}
        \sum_{q=1}^{3}
        \left(
        \widetilde{X}_{k+j}[r,q]
        -
        \overline{X}_{k+j}[r,q]
        \right)^2
        }.
        \label{eq:waypoint_discrepancy}
    \end{equation}
    The complete candidate state passes only if all eight waypoints
    satisfy the tolerance:
    \begin{equation}
        a_j
        =
        \mathbb{I}\!\left[
        \max_{r\in\{1,\ldots,8\}}d_{j,r}<\tau
        \right].
        \label{eq:verification_decision}
    \end{equation}
    Thus, a discrepancy violating the threshold at any waypoint
    rejects the entire candidate. This criterion limits local
    deviations from the target update without averaging them away
    across the trajectory. Since the comparison uses normalized
    action coordinates, $\tau$ is dimensionless.
    
    We then accept only the consecutive passing prefix in denoising
    order. Let $\ell$ denote its length. The accepted steps retain
    their draft states. At the first rejection, we adopt the
    target-corrected state and discard all subsequent candidates.
    The continuation state is
    \begin{equation}
        X_s^{\star}
        =
        \begin{cases}
            \overline{X}_{k+\ell+1}, & \ell<K_k,\\
            \widetilde{X}_{k+K_k}, & \ell=K_k,
        \end{cases}
        \label{eq:continuation_state}
    \end{equation}
    where $s=k+\min(\ell+1,K_k)$. The target alone completes any
    remaining denoising steps, with no further draft calls for
    that frame:
    \begin{equation}
        \widehat{X}_N
        =
        \mathcal{C}_{\theta}(X_s^{\star},s;c).
        \label{eq:final_action_state}
    \end{equation}
    Here, $\mathcal{C}_{\theta}$ follows the prescribed schedule
    from step $s$ to $N$ and is the identity when $s=N$.
    Accordingly, when all remaining steps pass verification,
    the terminal draft candidate is retained directly.
    The final state is denormalized to obtain the output trajectory.

%% file: sec/sections/04_experiments.tex
\section{Experiments}
\label{sec:experiments}

\subsection{Experimental setup}
\paragraph{Datasets and models: }We evaluate Trailblazer-F on SimWAM~\cite{zhao2026simwam} and
DriveWAM~\cite{shi2026drivewam} using NAVSIM
v1~\cite{dauner2024navsim}. Planning performance is evaluated on
all 12,146 scenes in the official \texttt{navtest} split.
For each target model, we collect its ten-step denoising state
sequences on \texttt{navtrain} and train a separate
drafter while keeping the target frozen. Checkpoints are selected
using a recording-disjoint validation split, and test scenes are
excluded from drafter fitting and checkpoint selection.
Across all experiments, Trailblazer-F uses a fixed local verification threshold of $\tau=0.05$ for both SimWAM and DriveWAM.
\paragraph{Hardware Configuration: }We measure inference efficiency on two platforms: a server
equipped with an NVIDIA L20 GPU and a vehicle-mounted NVIDIA
Jetson AGX Orin Developer Kit with 64\,GB of unified memory and
an integrated Ampere GPU. The Orin platform runs Ubuntu 22.04
with L4T R36.4.3 and CUDA 12.6.
We compare Trailblazer-F with each target's native ten-step
sampler, reporting end-to-end and action-flow latency alongside
the NAVSIM planning metrics.
\paragraph{Evaluation metrics: }
We evaluate planning quality using the official NAVSIM
v1 metrics~\cite{dauner2024navsim}: no at-fault collisions (NC),
drivable area compliance (DAC), time-to-collision within bound
(TTC), comfort (C), and ego progress (EP), together with their
aggregated Predictive Driver Model Score (PDMS).
\begin{table}[t]
\centering
\caption{Action-flow acceleration comparison.}
\label{tab:action_flow_acceleration}
\scriptsize
\setlength{\tabcolsep}{7pt}
\renewcommand{\arraystretch}{1.05}
\begin{tabular}{@{}lrrrrrr@{}}
\toprule
 & \multicolumn{3}{c}{SimWAM}
 & \multicolumn{3}{c}{DriveWAM} \\
\cmidrule(lr){2-4}\cmidrule(l){5-7}
Method
 & ms$\downarrow$ & $\times$$\uparrow$ & PDMS$\uparrow$
 & ms$\downarrow$ & $\times$$\uparrow$ & PDMS$\uparrow$ \\
\midrule
Native
 & 461.7 & 1.00 & 85.59
 & 1014.6 & 1.00 & 92.46 \\
BAC~\cite{ji2026bac}
 & 302.9 & 1.52 & 86.74
 & 820.3 & 1.24 & 92.45 \\
FastFlow~\cite{bajpai2026fastflow}
 & 269.7 & 1.71 & 84.46
 & 726.7 & 1.40 & 91.33 \\
TeaCache~\cite{liu2025teacache}
 & 234.5 & 1.97 & 86.26
 & 760.1 & 1.33 & 92.47 \\
\midrule
\textbf{Trailblazer-F}
 & \textbf{198.9} & \textbf{2.32} & 85.57
 & \textbf{653.7} & \textbf{1.55} & 92.45 \\
\bottomrule
\end{tabular}
\par\vspace{2pt}
\begin{minipage}{\columnwidth}
\scriptsize
\end{minipage}
\end{table}

\subsection{Main Results}
\label{sec:main_results}

\cref{tab:local_results} summarizes planning performance and
inference efficiency on NAVSIM v1. We compare our method with
the native ten-step sampler of each target model, while including
reported results from other driving methods to provide
planning-performance context.

\paragraph{Planning performance: }
Our method maintains planning scores close to those of the
original target models. SimWAM's PDMS changes from 86.04 to
86.08, while DriveWAM's PDMS changes from 90.25 to 90.17,
corresponding to differences of only $+0.04$ and $-0.08$
points, respectively. Across both backbones, all reported
planning subscore differences remain within 0.12 points
on the 0--100 scale.
These results show that the accelerated models retain
near-baseline planning performance across the evaluated
dimensions, supporting the effectiveness of target verification
in preserving planning quality during speculative inference.

\paragraph{Inference efficiency:}
Our method substantially reduces inference latency on the
NVIDIA L20 server. SimWAM's mean latency decreases from
460.3\,ms to 198.9\,ms, yielding a $2.31\times$ speedup
and a 56.8\% reduction. Notably, this brings SimWAM below
200\,ms per prediction, corresponding to approximately
5\,Hz at the measured model-inference level, compared
with approximately 2.2\,Hz for the serial baseline.
DriveWAM also benefits, with latency decreasing from
1015.9\,ms to 655.2\,ms, achieving a $1.55\times$ speedup
and a 35.5\% reduction.
Together with the near-baseline planning scores, these
results demonstrate an improved quality--latency tradeoff.
In particular, the sub-200\,ms SimWAM result provides
greater computational headroom for frequent replanning
and addresses a key latency bottleneck in deploying
action-flow world models.
\begin{figure}[!t]
\centering
\includegraphics[width=0.45\textwidth, trim=0 0.4cm 0cm 0cm, clip]{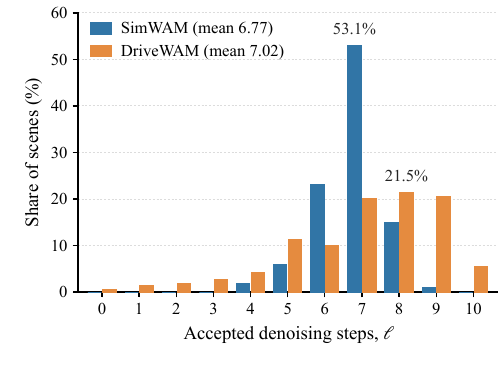}
\caption{Distribution of accepted denoising steps.}
\label{fig:dis_acc}
\end{figure}
\begin{figure*}[!t]
\centering
\includegraphics[width=1\textwidth, trim=0 0cm 0cm 0cm, clip]{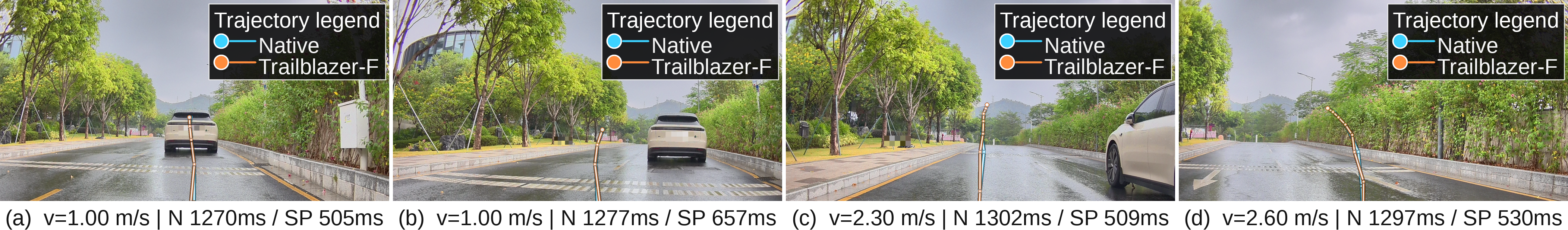}
\caption{\textbf{Trajectory comparison in real-vehicle experiments.}
Trailblazer-F (orange) produces trajectories closely aligned
with native sampling (cyan), while reducing flow latency in the illustrated scenes.}
\label{fig:vis_realword}
\end{figure*}
\paragraph{Comparison with Acceleration Baselines:}
\label{sec:acceleration_comparison}

We compare Trailblazer-F with BAC, FastFlow, and TeaCache
in \cref{tab:action_flow_acceleration}.
Trailblazer-F achieves the lowest inference latency among
the evaluated methods on both models, reaching 198.9\,ms
for SimWAM and 653.7\,ms for DriveWAM, corresponding to
$2.32\times$ and $1.55\times$ speedups over native sampling.
Meanwhile, its PDMS remains close to the corresponding
native baseline: 85.57 versus 85.59 for SimWAM, and
92.45 versus 92.46 for DriveWAM.
The comparison demonstrates that Trailblazer-F achieves
substantial acceleration while maintaining near-native
planning performance across both evaluated models.
\paragraph{Acceptance behavior:}
Figure~5 shows the distribution of accepted denoising steps across scenes.
SimWAM accepts an average of 6.77 out of 10 steps, with 53.1\% of scenes
accepting exactly seven steps. DriveWAM exhibits a broader distribution,
averaging 7.02 accepted steps and peaking at eight steps (21.5\%).
These results show that multi-step proposals frequently pass target
verification, allowing several sequential denoising updates to be
replaced by a single parallel verification round.
\subsection{Onboard Evaluation on NVIDIA Orin}
\label{sec:orin_evaluation}

\begin{table}[t]
    \centering
    \caption{Inference efficiency on NVIDIA Orin. 
    Latency: mean ± standard deviation (10 runs).
    }
    \label{tab:orin_latency}
    \small
    \setlength{\tabcolsep}{3pt}
    \renewcommand{\arraystretch}{1.05}
    \resizebox{\columnwidth}{!}{%
    \begin{tabular}{@{}lrrrr@{}}
        \toprule
        Method
        & \shortstack{E2E mean\\(ms) $\downarrow$}
        & \shortstack{Flow mean\\(ms) $\downarrow$}
        & \shortstack{Speedup\\$\uparrow$}
        & \shortstack{Accept.\\(steps/10)} \\
        \midrule
        \multicolumn{5}{@{}l}{\textit{SimWAM}} \\
        Native (10 steps)
        & $1709.6 \pm 49$
        & 1292.3
        & $1.00\times$
        & -- \\
        Trailblazer-F (Ours)
        & $\mathbf{870.2} \pm 149$
        & \textbf{451.8}
        & $\mathbf{1.96\times}$
        & 6.83 \\
        \midrule
        \multicolumn{5}{@{}l}{\textit{DriveWAM}} \\
        Native (10 steps)
        & $5105.3 \pm 262$
        & 4398.2
        & $1.00\times$
        & -- \\
        Trailblazer-F (Ours)
        & $\mathbf{3732.7} \pm 778$
        & \textbf{3026.6}
        & $\mathbf{1.37\times}$
        & 6.94 \\
        \bottomrule
    \end{tabular}%
    }
\end{table}

We evaluate inference efficiency on the vehicle-mounted NVIDIA
Jetson AGX Orin platform to assess the benefits on onboard hardware.
As shown in \cref{tab:orin_latency}, Trailblazer-F reduces
SimWAM's mean E2E latency from 1709.6\,ms to 870.2\,ms,
achieving a $1.96\times$ speedup and a 49.1\% reduction.
For DriveWAM, mean E2E latency decreases from 5105.3\,ms
to 3732.7\,ms, yielding a $1.37\times$ speedup and a 26.9\%
reduction.
The average accepted prefix lengths are 6.83 and 6.94 steps
out of the ten-step rollout for SimWAM and DriveWAM, respectively.

\begin{table*}[t]
  \centering
  \caption{\textbf{Role of target verification and correction.}}
  \label{tab:draft_only_ablation}
  \small
  \setlength{\tabcolsep}{4pt}
  \begin{tabular}{llrrrrrrrr}
    \toprule
    Model & Method & NC $\uparrow$ & DAC $\uparrow$ & TTC $\uparrow$
    & C. $\uparrow$ & EP $\uparrow$ & PDMS $\uparrow$
    & \shortstack{Inference\\(ms) $\downarrow$} & \shortstack{Accept\\(steps/10)} \\
    \midrule
    SimWAM & Serial target (10 steps) & 97.77 & 94.11 & 93.69 & 100.00 & 81.50 & 86.04 & 460.3 & -- \\
     & Draft-only (one-shot) & 67.69 & 66.05 & 57.30 & 64.79 & 35.62 & 33.71 & 73.7 & -- \\
     & Draft-only (recursive, 10 calls) & 67.05 & 63.94 & 55.13 & 64.80 & 34.92 & 32.47 & 78.2 & -- \\
     & Trailblazer-F & 97.76 & 94.22 & 93.57 & 99.99 & 81.58 & 86.08 & 198.9 & 6.8 \\
    \midrule
    DriveWAM & Serial target (10 steps) & 98.33 & 98.15 & 95.40 & 99.99 & 84.33 & 90.25 & 1015.9 & -- \\
     & Draft-only (one-shot) & 75.08 & 72.78 & 64.00 & 99.95 & 42.99 & 44.96 & 374.2 & -- \\
     & Draft-only (recursive, 10 calls) & 75.50 & 72.89 & 64.36 & 99.93 & 43.33 & 45.35 & 377.9 & -- \\
     & Trailblazer-F & 98.31 & 98.11 & 95.32 & 99.99 & 84.28 & 90.17 & 655.2 & 7.0 \\
    \bottomrule
  \end{tabular}
\end{table*}

The flow latency measurements further locate the source of
these gains.
SimWAM's mean flow latency decreases from 1292.3\,ms to
451.8\,ms, a 65.0\% reduction, while DriveWAM's decreases
from 4398.2\,ms to 3026.6\,ms, a 31.2\% reduction.
For both models, the flow latency reductions closely match
the E2E savings, indicating that the gains primarily arise
within the flow computation targeted by our method.
In our deployment, DriveWAM's video-flow computation becomes
more time-consuming on Orin, contributing to its higher
absolute E2E latency.
Nevertheless, Trailblazer-F accelerates both models onboard, extending its benefits beyond server GPUs.

\subsection{Real-Vehicle Experiments}
\label{sec:real_vehicle}

We deploy SimWAM on a real vehicle and evaluate Trailblazer-F
in a representative lane-change overtaking scenario.
As shown in Fig.~6, the ego vehicle approaches a vehicle ahead,
moves into the adjacent lane, and passes it.
We compare the trajectories produced by native sampling and
Trailblazer-F at four stages of this maneuver.

The trajectories generated by Trailblazer-F (orange) closely
follow those of native sampling (cyan), including during the
lateral transition and passing stages.
Across the four illustrated snapshots, flow latency ranges from 1270 to 1302 ms for native sampling and from 505 to 657 ms for Trailblazer-F.
These observations show that Trailblazer-F reduces on-vehicle flow latency while retaining close agreement with the native planner’s trajectories in this overtaking scenario.
\subsection{Ablation Studies on Draft Model}
\label{sec:ablations}

We investigate whether the lightweight drafter can replace
the target model during action generation.
For each backbone, we reuse the same draft checkpoint as
the full method and evaluate two variants without target
verification or correction.
The \emph{one-shot} variant directly outputs the terminal
candidate from a single draft forward pass.
The \emph{recursive} variant feeds each predicted next-step
state back into the drafter, performing ten successive
calls to complete the denoising sequence.

As shown in \cref{tab:draft_only_ablation}, draft-only
generation substantially reduces inference latency but
severely degrades planning performance.
For SimWAM, one-shot generation reduces latency to
73.7\,ms, while PDMS drops from 86.04 to 33.71.
For DriveWAM, latency decreases to 374.2\,ms, but PDMS
falls from 90.25 to 44.96.
These degradations extend across collision avoidance,
drivable-area compliance, time-to-collision, and ego
progress. Recursive generation does not recover the
lost performance, yielding PDMS values of 32.47 and
45.35 for the two backbones, respectively.

In contrast, Trailblazer-F combines draft proposals with
target verification and correction, achieving PDMS values
of 86.08 and 90.17 while reducing inference latency to
198.9\,ms and 655.2\,ms.
These results show that the trained drafter is effective
as a proposal mechanism, but is insufficient as a
standalone replacement for the target.
Target-guided verification and correction enable the
full method to exploit inexpensive proposals while
maintaining planning scores close to those of the
native sampler.

%% file: sec/sections/05_conclusion.tex
\section{Conclusion}

We presented Trailblazer-F, a speculative action-flow inference
framework that reduces inference latency while maintaining planning
performance close to native sampling.
Its lightweight drafter proposes multiple future states for parallel
target verification, preserving the target model's parameters and
architecture.
Beyond the demonstrated speedups, this modular design provides
flexibility to complement other efficiency improvements, including
WAM distillation and visual pruning.
By adapting the drafter to an optimized target, our framework can
explore further acceleration through proposals verified against
that target's own dynamics.
Future work will investigate these combinations and evaluate their
cumulative efficiency gains while preserving the planning capability
of the deployed target.
\label{sec:conclusion}

%% file: sec/sections/06_llm_usage.tex
\section*{LLM Usage Statement}
We used OpenAI Codex to polish the manuscript text, assist with code debugging, and refine figure layouts. 